\documentclass[runningheads]{llncs}

\usepackage{eccv}

\usepackage{eccvabbrv}

\usepackage{graphicx}
\usepackage{booktabs}
\usepackage[accsupp]{axessibility}  % Improves PDF readability for those with disabilities.

\usepackage{hyperref}

\usepackage{orcidlink}

\usepackage{graphics}
\usepackage{array}
\usepackage{multirow}
\usepackage{wrapfig}
\usepackage{hhline}
\usepackage{pifont}% http://ctan.org/pkg/pifont
\newcommand{\topscore}[1]{\textcolor{blue}{\textbf{#1}}} % vector notation
\newcolumntype{K}[1]{>{\centering\arraybackslash}p{#1}}
\usepackage{duckuments}
\usepackage{tikzducks}
\usepackage{graphicx}
\usepackage{adjustbox}
\usepackage{pifont}
\usepackage{float}

\usepackage{atbegshi}
\newcommand{\placetextbox}[3]{
  \setbox0=\hbox{#3}% Put <stuff> in a box
  \AtBeginShipoutNext{\AtBeginShipoutUpperLeft{%
    \put(\dimexpr#1\paperwidth\relax,-\dimexpr#2\paperheight\relax)
    {\vtop{{\null}\makebox[0pt][c]{#3}}}%
  }}%
}

\newcommand\blfootnote[1]{%
  \begingroup
  \renewcommand\thefootnote{}\footnote{#1}%
  \addtocounter{footnote}{-1}%
  \endgroup
}

\begin{document}

\title{Intrinsic Single-Image HDR Reconstruction} 

%\titlerunning{Intrinsic SI-HDR}

\author{Sebastian Dille$^*$~\orcidlink{0000-0003-0390-2803} \and
Chris Careaga$^*$~\orcidlink{0000-0002-0800-1118} \and
Ya\u{g}{\i}z Aksoy~\orcidlink{0000-0002-1495-0491}}

\authorrunning{S.~Dille et al.}

\institute{Simon Fraser University\\ Burnaby, BC V5A 1S6, Canada\\
\email{sdille@sfu.ca}
\blfootnote{($^{*}$) denotes equal contribution.}}

\maketitle

\placetextbox{0.14}{0.03}{\includegraphics[width=4cm]{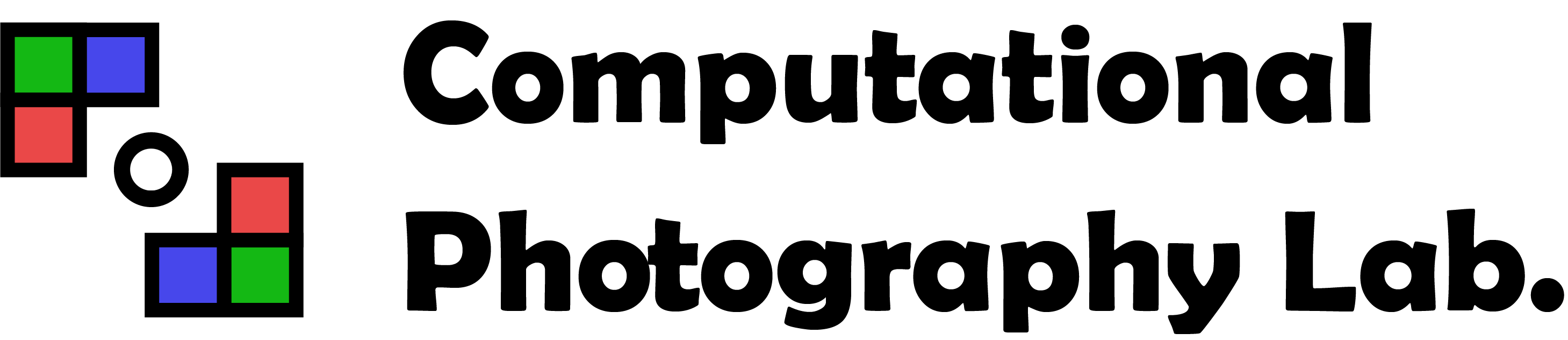}}
\placetextbox{0.8}{0.03}{Find the project web page here:}
\placetextbox{0.8}{0.045}{\textcolor{purple}{\url{https://yaksoy.github.io/intrinsicHDR/}}}

\begin{center}
    \centering
    \includegraphics[width=0.95\linewidth]{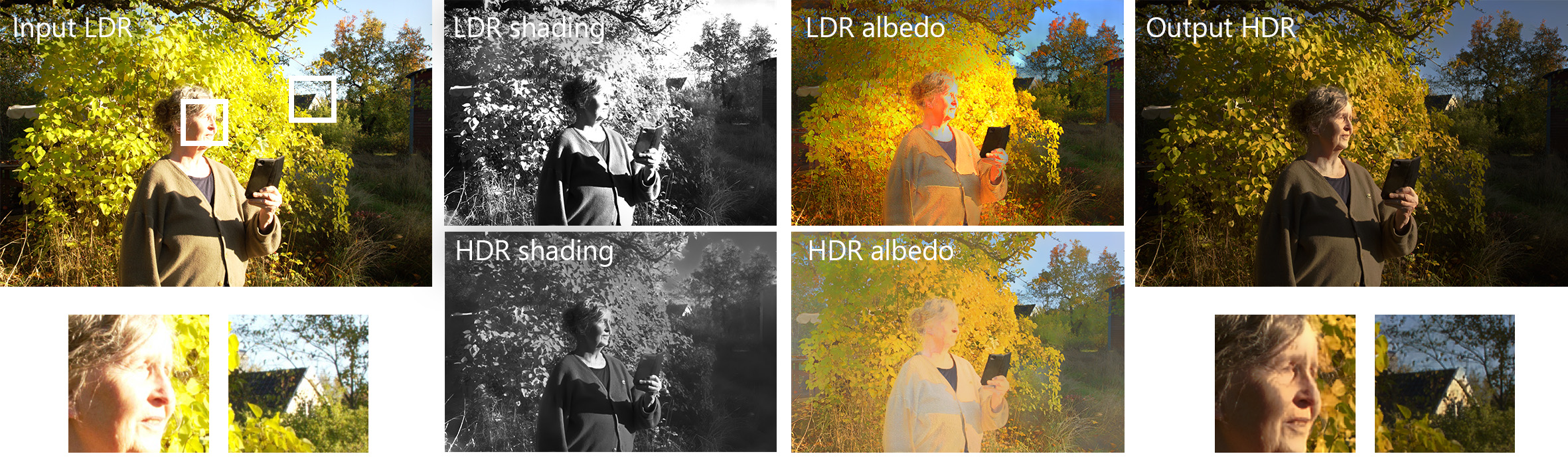}\\
    \includegraphics[width=0.95\linewidth]{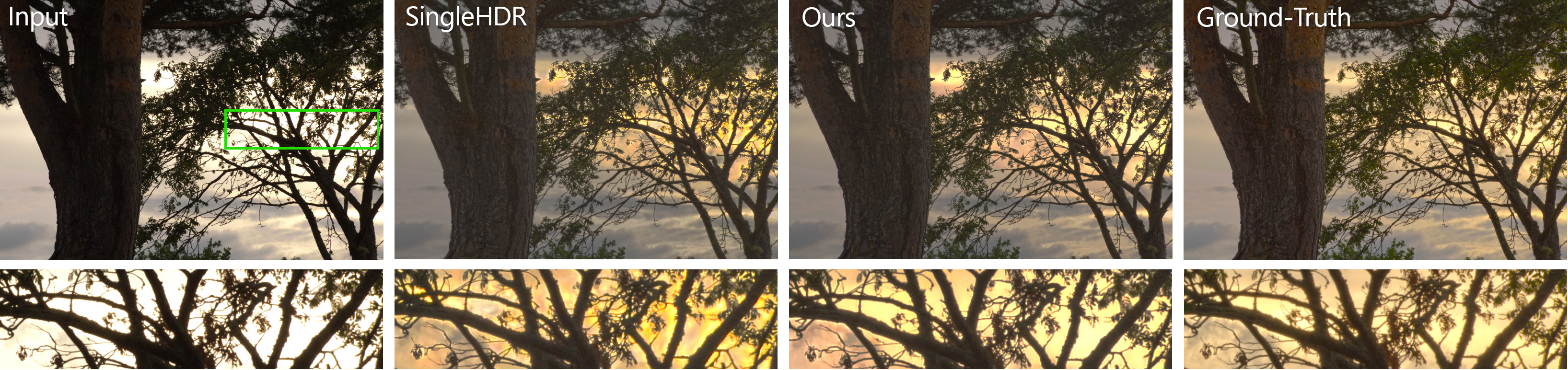}
    \captionof{figure}{
      In this work, we propose a physically-inspired HDR reconstruction pipeline by conducting reconstruction individually on intrinsic components. 
      Our final reconstruction, as a result, is effective in recovering the high-luminance details and the lost colors.
    }
    \label{fig:teaser}
\end{center}%

\begin{abstract}
    The low dynamic range (LDR) of common cameras fails to capture the rich contrast in natural scenes, resulting in loss of color and details in saturated pixels. 
    Reconstructing the high dynamic range (HDR) of luminance present in the scene from single LDR photographs is an important task with many applications in computational photography and realistic display of images.
    The HDR reconstruction task aims to infer the lost details using the context present in the scene, requiring neural networks to understand high-level geometric and illumination cues. 
    This makes it challenging for data-driven algorithms to generate accurate and high-resolution results.
    In this work, we introduce a physically-inspired remodeling of the HDR reconstruction problem in the intrinsic domain. 
    The intrinsic model allows us to train separate networks to extend the dynamic range in the shading domain and to recover lost color details in the albedo domain. 
    We show that dividing the problem into two simpler sub-tasks improves performance in a wide variety of photographs.

\end{abstract}

\section{Introduction}
\label{sec:intro}

Real-world scenes contain a wider range of luminance than most consumer-grade cameras can record or store. 
As a result, everyday photographs can only represent a low dynamic range (LDR) version of the scene that lacks details and color in very dark or bright regions. 
High dynamic range (HDR) imaging aims to capture and store the full contrast present in the scene. 
HDR images allow realistic display of photographs using high-contrast displays as well as many computational photography applications such as tone mapping and illumination-aware image editing. 
However, HDR imaging typically requires multi-exposure image stacks which limits their applications to static scenes.

To extend the HDR-enabled applications to everyday photographs, there has been a growing interest in methods to estimate HDR images from only LDR photographs called single-image HDR reconstruction, or LDR-to-HDR conversion. 
Single-image HDR reconstruction aims to extend the dynamic range of LDR photographs by recovering the details lost during capture.
The missing information in LDR images is typically characterized by saturated pixels in bright regions, lacking textural details and color. 
This makes HDR reconstruction a high-level problem, requiring networks to utilize geometric and context-dependent cues in the scene to infer the lost information.
Formulating this reconstruction problem using standard color spaces such as RGB or Luv requires the networks to statistically model the complex relationship between luminance, surface colors, and geometry, which limits their performance.

In this work, we present a physically-inspired remodeling of the HDR reconstruction problem in the intrinsic domain. 
Intrinsic image decomposition is a mid-level vision problem that separates the effect of the illumination in the scene from the reflectance of surfaces~\cite{careaga2023intrinsic}, representing the input as a combination of a \emph{shading} and an \emph{albedo} layer. 
The shading of an image is highly correlated with the scene geometry and illumination, while the albedo contains the color information and the high-frequency textures. 
The saturated pixels we aim to recover create different degradations in the shading and albedo layers. 
The clipping of high values is observed in the shading, while the loss of color information causes desaturation in the albedo. 
We propose an HDR reconstruction pipeline that individually addresses the loss of information in the LDR image in the two intrinsic components. 
Our HDR shading reconstruction network extends the dynamic range, taking advantage of the high correlation between shading and the scene geometry. 
Our LDR albedo reconstruction network, on the other hand, recovers the lost color information in saturated regions. 
A final refinement stage combines the two components to generate our full HDR reconstruction. 
Figure~\ref{fig:teaser} shows different stages of our pipeline.

In our formulation, we take advantage of the intrinsic image formation model by dividing the recovery of lost details and colors into two separate tasks which are more convenient for networks to model. 
We show that our physically inspired approach can generate high-fidelity details with rich colors in complex in-the-wild scenes. 
We evaluate our method qualitatively and quantitatively to demonstrate the state-of-the-art performance of our formulation in a wide variety of scenes.

\section{Related work}
\label{sec:related}
Common imaging sensors for photography are inherently limited in the portion of the scene's dynamic range they can capture in a single shot. This results in saturated pixels in bright regions, representing a significant challenge in the capture and realistic display of real-world scenes.

Multi-exposure fusion methods can reconstruct an HDR image by combining a stack of LDR images taken at different exposures~\cite{mann1995being,debevec1997recovering,zhang2023revisiting}. Their formulation successfully addresses contrast range extension and reversal of the quantization and nonlinearity of LDR images.
The multi-exposure capture, however, assumes a static camera and scene, requiring a challenging alignment process and limiting their application scenario. 
To enable HDR-related applications in everyday photography, single-image HDR reconstruction approaches aim to recover the lost information directly from an LDR image. 
Earlier approaches apply specially designed filters \cite{rempel2007ldr2hdr,kovaleski2014high} or median-cut \cite{banterle2006inverse,debevec2005median} to fill in overexposed regions, using different heuristics to scale the intensity of a selected area. 
Similarly, Bayesian \cite{zhang2004estimation} and graph-based formulations \cite{xu2010correction} have been introduced for chroma restoration.
Due to the high-level nature of the reconstruction problem, these low-level approaches struggle to generate realistic details in real-world scenarios.

In their concurrent publications, Eilertsen~\etal~\cite{eilertsenHDRImageReconstruction2017} and Endo~\etal~\cite{endoDeepReverseTone2017} present the first data-driven approaches to the single-image HDR reconstruction problem. 
They train their networks for direct conversion to HDR from the LDR input, addressing quantization, nonlinearity, and saturation in the LDR image in a single step. 
Many works in recent literature adapt this unified inverse model~\cite{liu2020single} and develop tailored solutions for underexposed~\cite{chen2021hdrunet,guo2022lhdr} or overexposed~\cite{santosSingleImageHDR2020} areas or losses for color recovery \cite{marnerides2018expandnet}. 
The direct estimation of HDR outputs, however, creates the challenge of an unconstrained color distribution that hinders training efficiency. Several methods \cite{endoDeepReverseTone2017, leeDeepRecursiveHDRI2018, kim2021end, le2023single, zhang2023revisiting, cevr_2023} instead adopt the concept of exposure stacks and predict a set of over- and underexposed images from the input LDR image. With each prediction still in the LDR domain, the HDR image can later be created with multi-exposure fusion~\cite{debevec1997recovering}. Other methods~\cite{zhang2021deep,wang2023glowgan} employ generative approaches, either generating separate LDR detail layers~\cite{zhang2021deep} or full HDR images directly~\cite{wang2023glowgan}, with image-specific HDR reconstruction via GAN inversion.  
Liu~\etal~\cite{liu2020single} approach HDR reconstruction by inverting the in-camera image formation process, modeling the dequantization, nonlinearization, and saturated pixel reconstruction in three separate steps. They show that this decomposition of the task simplifies the modeling of each individual sub-problem.

In this work, we adopt the modular formulation of Liu~\etal~\cite{liu2020single} and focus on recovering the lost details and color information in the saturated pixels. Our method takes inspiration from prior works that formulate computational photography problems in the intrinsic image domain~\cite{Maralan2023Flash,careagaCompositing}. Rather than defining a direct mapping from LDR to HDR, we divide this task into two sub-problems: shading reconstruction that extends the dynamic range, and reflectance reconstruction that recovers the color information. Our physical deconstruction of the problem significantly improves the recovery of details and color information.%

\section{Background and Motivation}
\label{sec:background}

The focus of our work is the formulation of the single-image LDR-to-HDR reconstruction problem in the intrinsic domain. 
We begin our discussion by first introducing the LDR imaging pipeline and the intrinsic model of image formation. 
We then draw connections between the two models and motivate our intrinsic HDR formulation.

\subsection{LDR image formation and HDR reconstruction}

\begin{figure}[t]
    \centering
    \includegraphics[width=\linewidth]{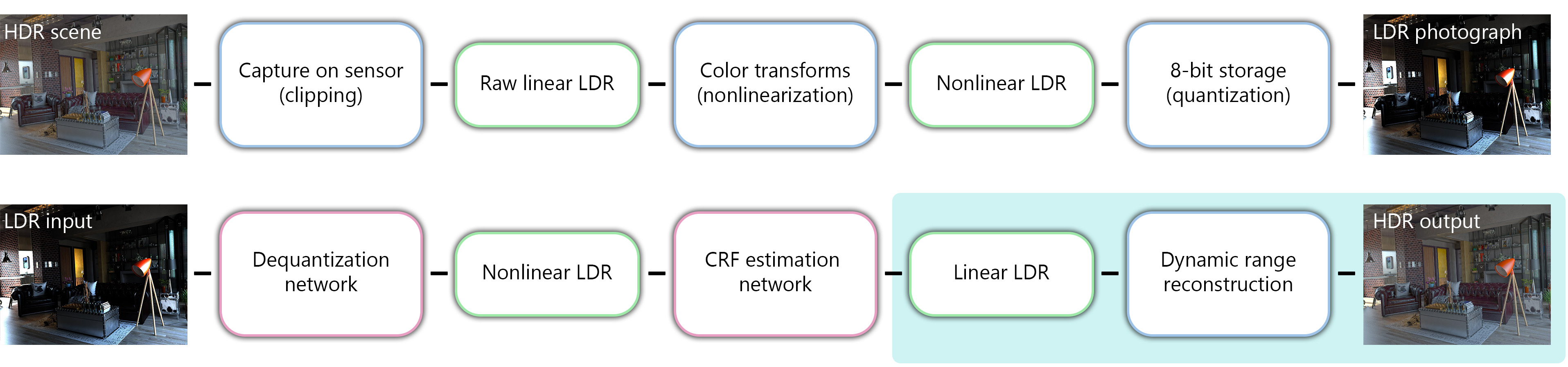}
    \caption{The in-camera pipeline modeling LDR image formation is shown at the top. Liu~\etal~\cite{liu2020single} propose the inverse pipeline shown on the bottom, addressing each type of information loss individually in three steps. 
    In this work, we use the pre-trained networks of Liu~\etal~\cite{liu2020single} for the dequantization and linearization tasks (shown in pink) and focus on developing an intrinsic model for the dynamic range reconstruction problem highlighted in teal.}
    \label{fig:LDRpipeline}
\end{figure}

The acquisition of standard RGB photographs involves a series of operations in the camera-internal image processing pipeline, or ISP, that result in different forms of information loss. 
The initial acquisition of the image is in a raw format, where the luminance received by the sensor from a real-world scene is scaled and stored in 16- or 32-bit data structures. 
The scale of the captured raw values is determined by the analog camera settings such as the shutter speed, the aperture size, and the ISO. 
Given the natural high dynamic range of real-world scenes, the raw values are clipped at a maximum due to saturation at the sensor level. 
As shown in Figure~\ref{fig:LDRpipeline}, the remainder of the ISP applies nonlinear color and gamma corrections before finally decreasing the bit rate to 8 bits per channel for storage.

The LDR image formation process causes three different types of corruption of the HDR information: quantization, non-linearization, and clipping. Single-image HDR reconstruction methods, as a result, have to address all three of these challenges. 
While many methods in the literature address all three simultaneously within a single architecture, Liu~\etal~\cite{liu2020single} design their pipeline to address each type of information loss with a separate dedicated network. 
They propose a dequantization network to first increase the bit depth of the input LDR image and then estimate a CRF curve for linearization. 
Both tasks are convenient for a network to learn, taking advantage of the ease of dataset generation for dequantization and a simple parametric curve estimation for linearization. 

They then reconstruct the values lost to clipping with a third network in what they call the hallucination stage. 
This last stage of the inverse pipeline addresses the most significant challenge in single-image HDR reconstruction, namely the recovery of the highlights and the color information in clipped regions.  
Diverging from Liu~\etal's terminology, we refer to this stage as dynamic range extension. 
We adopt Liu~\etal's inverse pipeline and use their pre-trained networks for dequantization and linearization. 
In the following, we focus on the dynamic range extension problem and present our solution using an intrinsic model of image formation.

\begin{figure}[t]
    \centering
    \includegraphics[width=\linewidth]{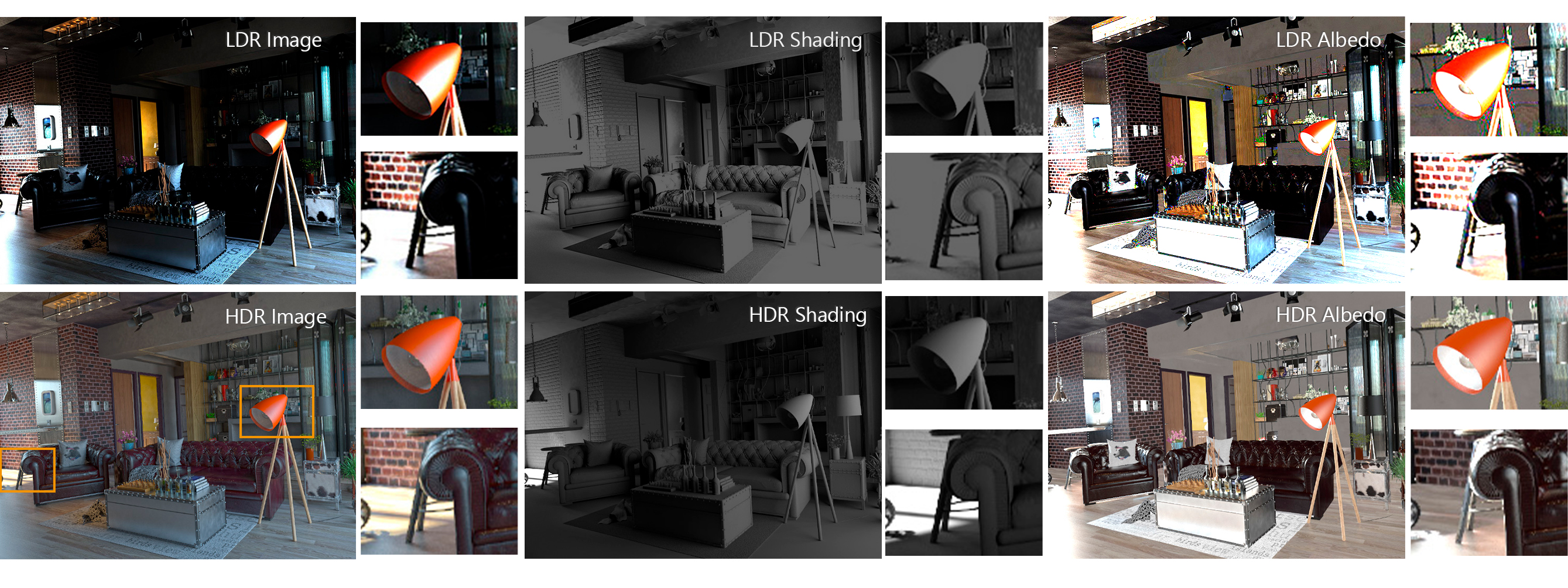}
    \caption{We show the different effects of clipping on the albedo and the shading for a tonemapped HDR image and its LDR counterpart. When clipping the image, the LDR shading loses information compared to HDR. We scale the intensity of both shadings to show the effect. At the same time, the LDR albedo becomes saturated.}
    \label{fig:IntrinsicHDR}
\end{figure}

\subsection{Intrinsic image decomposition}
Intrinsic image decomposition is a fundamental mid-level computer vision problem that represents a core step in inverse rendering pipelines~\cite{barrow1978recovering, garces2022survey}. 
It aims to separate the image into illumination and reflectance components following the physical image formation model. 
Adopting the commonly used Lambertian shading assumption with a single-channel shading representation \cite{careaga2023intrinsic}, the intrinsic equation can be written as:
\begin{equation}
    I_{rgb}  = A_{rgb} \cdot S,
\label{eq:intrinsicmodel}
\end{equation}
where $I$ represents the linear image reflecting relative luminance, $A$ represents albedo, i.e. the surface reflectance containing the color information, and $S$ represents shading reflecting the effect of illumination in the scene.

Shading and albedo represent two orthogonal characteristics of a scene, one containing the effect of light and its interaction with the scene geometry and the other representing material properties. 
As a result, the two intrinsic components have very different characteristics as demonstrated in Figure~\ref{fig:IntrinsicHDR}. 
Shading is highly correlated with surface orientations, appearing smooth over continuous surfaces and having a high gradient in depth discontinuities and shadow boundaries. 
Albedo, on the other hand, contains the high-frequency textures present on the materials in the scene while having a sparse set of values per image as the variation in brightness due to geometry is isolated in the shading layer. 
In the single-channel shading representation, albedo also contains any variation caused by colorful illumination~\cite{careaga2023intrinsic}.

\subsection{Intrinsics in LDR and HDR}

The bounded representation of LDR images limits the range of luminance values they can contain.
HDR reconstruction aims to extend this range of luminance by exploiting statistical properties of real-world photographs derived from high-level scene properties such as geometry and appearance. 

A brief look at the intrinsic components of HDR images, as shown in Figure~\ref{fig:IntrinsicHDR}, reveals the different characteristics of the shading and albedo layers. 
The distribution of reflectance values in real-world scenes typically has a narrow distribution, making the albedo a variable that can be represented linearly between $[0,1]$. 
The high dynamic range of the HDR image, on the other hand, is reflected in the shading layer that similarly has an unbounded representation. 

As a result, the limited dynamic range of LDR images has different effects on the shading and albedo layers. 
As Figure~\ref{fig:IntrinsicHDR} demonstrates, the clipping of HDR values results in a similar clipping in shading, showing the relationship between the dynamic range and illumination that has been previously discussed in the literature~\cite{eilertsenHDRImageReconstruction2017}. 
The loss of colors induced by clipping, on the other hand, appears in the albedo layer as the dullification of colors in over-exposed regions. 

Based on this observation, we formulate the HDR reconstruction problem separately in shading and albedo. 
In our setup, the network tasked with converting LDR shading into HDR shading models the high dynamic range of the illumination in the scene, taking advantage of the high correlation between shading and high-level scene properties such as geometry. 
We task another network with converting the LDR albedo into HDR albedo, training it to recover the lost color information in clipped regions. 
We detail our full pipeline in Section~\ref{sec:method}.

Our approach effectively separates the two challenges in dynamic range reconstruction, reconstruction of high luminance regions and recovery of the color information, into two separate tasks. 
Formulating the reconstruction of high luminance in the shading domain results in a more convenient way for the network to statistically model the problem. This can be attributed to the high correlation between shading and geometry, which is often violated in luminance due to textured surfaces as highlighted in Figure~\ref{fig:LUVvsIntrinsic}.
Formulating the color recovery problem in the albedo domain simplifies the task similarly, allowing the exploitation of the sparsity of reflectance values in real-world scenes. 
Our albedo LDR-to-HDR network, as a result, acts comparable to a simple inpainting model, filling in the lost color details based on the scene content.

\section{Dynamic Range Extension via Intrinsics}
\label{sec:method}

\begin{figure}[t]
    \centering
    \includegraphics[width=\linewidth]{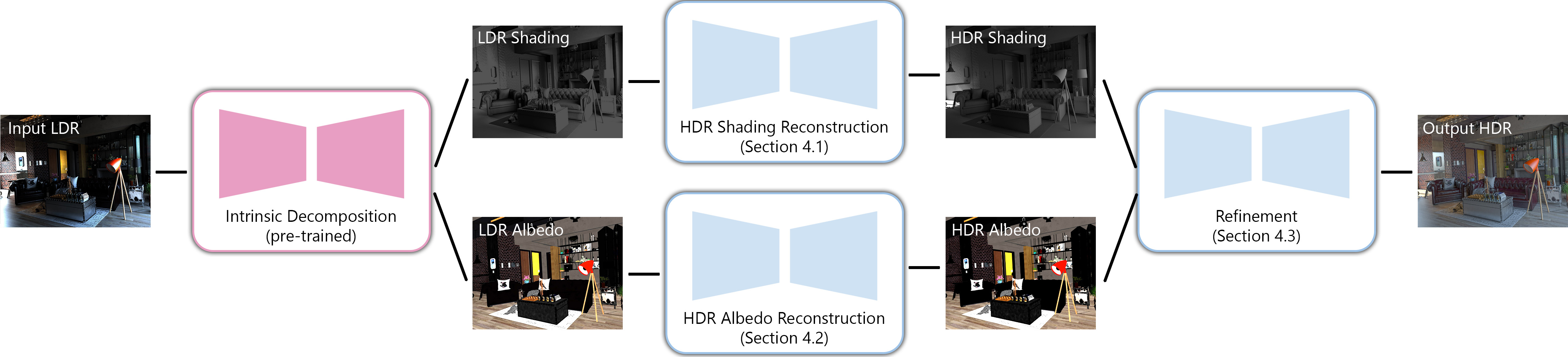}
    \caption{Our HDR reconstruction pipeline starts with the intrinsic decomposition of the image using an off-the-shelf method \cite{careaga2023intrinsic}. The LDR reconstruction is then done individually on shading and albedo layers, which are then combined in a final refinement stage. The input LDR image is provided to each network shown in blue.}
    \label{fig:IntrinsicPipeline}
\end{figure}

In this work, we propose a physically inspired system for dynamic range extension in the single-image HDR reconstruction pipeline.
The input to our pipeline is the dequantized and linearized LDR photograph $I_L$. 
We first compute the intrinsic components of $I_L$, the LDR shading $S_L$, and albedo $A_L$, using the in-the-wild intrinsic decomposition method by Careaga and Aksoy~\cite{careaga2023intrinsic}. 
We then estimate an HDR shading and an HDR albedo from the corresponding LDR components using dedicated networks. 
The estimated HDR intrinsic components are then combined and refined by a subsequent HDR reconstruction network yielding our final result. 
We present an overview of our pipeline in Figure~\ref{fig:IntrinsicPipeline}. 
In this section, we present our formulation and training details for each step.

\subsection{HDR shading reconstruction}

The shading layer in the intrinsic model contains the effect of illumination in the scene. 
The high dynamic range of illumination in the scene, shadows, and specular objects cause a long-tailed distribution of values in the shading map. 
An LDR input with clipped values similarly produces a clipped LDR shading layer after intrinsic decomposition. 
We train our HDR shading reconstruction network to extend the range of shading values beyond this clipping.

\begin{figure}[t]
    \centering
    \includegraphics[width=\linewidth]{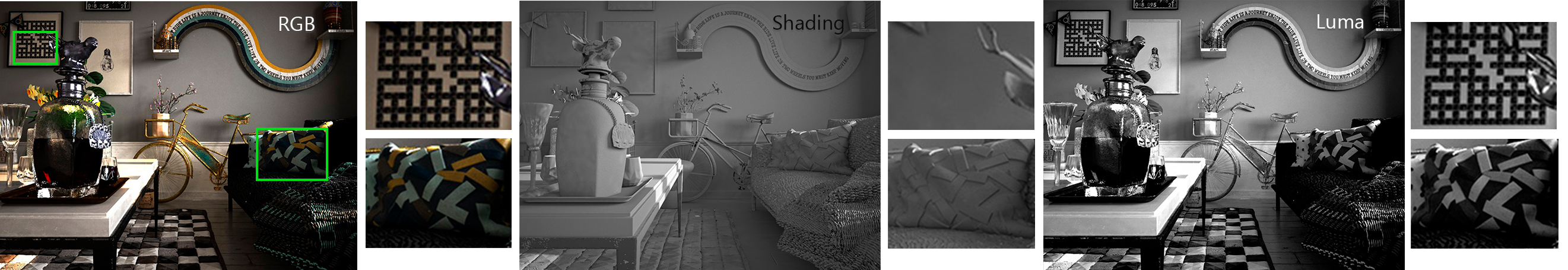}
    \caption{We show the difference between the shading representation and the L-channel from LUV. The shading correlates well with the geometry, while the L-channel includes texture information as shown in the green insets.}
    \label{fig:LUVvsIntrinsic}
\end{figure}

Several works in the literature define the dynamic range extension task in the luminance domain, targeting a similar goal of isolating the effect of illumination. 
In their work, Eilertsen~\etal~\cite{eilertsenHDRImageReconstruction2017} motivate this choice from a perceptual standpoint, citing earlier work that explores the effect of illumination versus surface colors in human perception~\cite{GilchristIlluminationPerception1984}.
The luminance, however, is computed as a linear combination of RGB channels and, as a result, still contains information on the reflectance. 
Shading, on the other hand, directly reflects the illumination and is accordingly highly correlated with the scene geometry as Figure~\ref{fig:LUVvsIntrinsic} demonstrates. 
Defining the range extension task in shading instead of luminance benefits the network to model the clipped shading values by easily utilizing the geometry-related cues in the input.

Following Careaga and Aksoy~\cite{careaga2023intrinsic}, we represent shading in the inverse domain that fits the long-tailed shading in a representation bounded in $[0,1]$:
\begin{equation}
    D_L = \frac{1}{S_L+1}. 
\end{equation}
The motivation behind this inverse representation is similar to the use of the log space in HDR reconstruction: It allows the definition of balanced loss functions on variables that could take very large or small values otherwise. 
The advantage of the inverse domain over the log space lies in its bounded nature that allows using the variable as input without clipping or normalization as well as the use of bounded activation functions such as sigmoid.

Our HDR shading reconstruction network receives the LDR image $I_L$ and the estimated LDR inverse shading $D_L$ as a concatenated $h \times w \times 4$ input. 
Its output is defined as the HDR inverse shading $D_H$.
We use multiple loss functions to train our network. 
The first one is the standard mean-squared error loss defined in the inverse space:
\begin{equation}
    \mathcal{L}_{mse}^{D} = MSE \left( D^*_H, D_H \right),
\end{equation}
where $D^*_H$ and $D_H$ denote the ground-truth and estimated HDR inverse shading, respectively. 
We also use the multi-scale gradient loss~\cite{li2018learning} as an edge-aware smoothness loss:
\begin{equation}
    \mathcal{L}_{msg}^D = MSG \left( D^*_H, D_H \right) = \frac{1}{N} \sum_m SAD \left( \nabla D^{*m}_H, \nabla D^m_H \right),
\end{equation}
where $\nabla$ denotes the spatial gradient and $D_H^m$ denotes the estimated inverse HDR shading at scale $m$.

Although the output of our HDR shading reconstruction network is defined only as the HDR inverse shading, we can use the ground-truth HDR image $I^*_H$ and the intrinsic model in Eq.~\ref{eq:intrinsicmodel} to compute an \emph{implied} HDR albedo $\hat{A}_H$~\cite{careaga2023intrinsic}:
\begin{equation}
    \hat{A}_H = \frac{I^*_H}{S_H}, \quad S_H = \frac{1 - D_H}{D_H}. 
\label{eq:impliedAlbedo}
\end{equation}
We define the MSE and multi-scale gradient loss also for the implied albedo:
\begin{equation}
    \mathcal{L}_{mse}^{\hat{A}} = MSE \left( A^*_H, \hat{A}_H \right), \quad \mathcal{L}_{msg}^{\hat{A}} = MSG \left( A^*_H, \hat{A}_H \right),
\end{equation}
and backpropagate these losses to our network through Eq.~\ref{eq:impliedAlbedo}. 

Our HDR shading reconstruction network only estimates the HDR shading and the HDR albedo component will be estimated with a separate network defined in the next section. 
Estimating the HDR intrinsic components separately results in possible reconstruction errors when these components are combined to produce an estimation for the HDR image. 
While we will refine the HDR reconstruction in our final stage with a dedicated network, we define these losses on the implied albedo to promote a better fit within the intrinsic model.

Our final loss for the HDR shading reconstruction network is defined as:
\begin{equation}
    \mathcal{L}^D = \mathcal{L}_{mse}^{D} + \gamma \cdot \mathcal{L}_{msg}^D + \mathcal{L}_{mse}^{\hat{A}} + \gamma \cdot \mathcal{L}_{msg}^{\hat{A}}, \quad \gamma = 0.1.
\end{equation}

\subsection{HDR albedo reconstruction}

The clipping in the LDR image formation process results in the loss of color information in the saturated pixels of the LDR albedo map. 
Our HDR albedo reconstruction network is tasked with recovering the corrupted colors. 

While the LDR shading map has saturated values, signaling the regions to be reconstructed to our HDR shading reconstruction networks, the LDR albedo map does not carry the same signal as albedo maps are not saturated. 
To denote the regions that require color reconstruction, we provide the network with a soft guidance mask $\alpha$ defined as:
\begin{equation}
    \alpha = \frac{max(0, I_L - \lambda)}{1 - \lambda}, \quad \lambda = 0.8,
\end{equation}
similar to the soft mask defined by Liu~\etal~\cite{liu2020single}. 
It should be noted that, unlike Liu~\etal, we use this mask as an input to our network rather than a blending mask. 

Our HDR albedo reconstruction network receives the LDR image $I_L$, the estimated LDR albedo $A_L$, and the soft attention mask $\alpha$, as a concatenated $h \times w \times 7$ input. 
Similar to our loss formulation for shading, we define the MSE and multi-scale gradient losses:
\begin{equation}
    \mathcal{L}_{mse}^{A} = MSE \left( A^*_H, A_H \right), \quad
    \mathcal{L}_{msg}^A = MSG \left( A^*_H, A_H \right),
\end{equation}
as well as losses on implied inverse shading $\hat{D}_H  = {A_H}/({I^*_H + A_H})$ using the ground-truth HDR image $I^*_H$ and the intrinsic model in Eq.~\ref{eq:intrinsicmodel}:
\begin{equation}
    \mathcal{L}_{mse}^{\hat{D}} = MSE \left( D^*_H, \hat{D}_H \right), \quad 
    \mathcal{L}_{msg}^{\hat{D}} = MSG \left( D^*_H, \hat{D}_H \right).
\end{equation}
Our final loss for the HDR albedo reconstruction network is defined as:
\begin{equation}
    \mathcal{L}^A = \mathcal{L}_{mse}^{A} + \gamma \cdot \mathcal{L}_{msg}^A + \mathcal{L}_{mse}^{\hat{D}} + \gamma \cdot \mathcal{L}_{msg}^{\hat{D}}, \quad \gamma = 0.1.
\end{equation}

\begin{figure}[t]
    \centering
    \includegraphics[width=\linewidth]{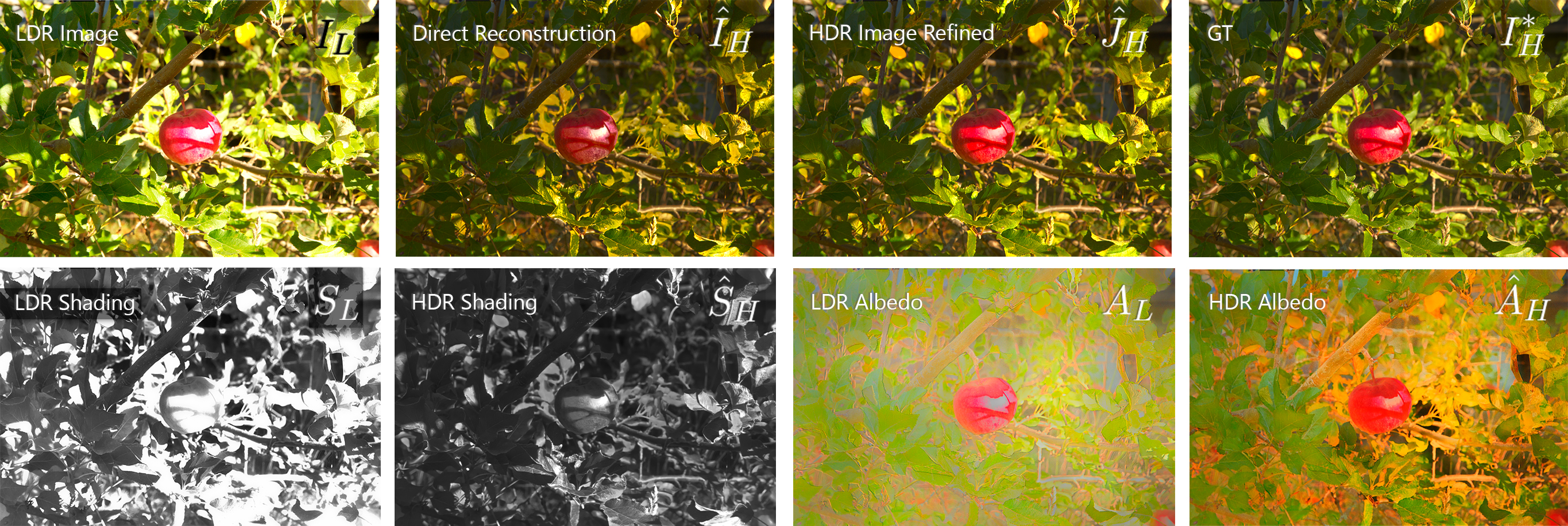}
    \caption{We show the intermediate results of our pipeline from an LDR input compared with its ground truth. The direct HDR reconstruction from the estimated HDR intrinsics can lead to small artifacts which are reduced by our refinement network.}
    \label{fig:PipelineIntermediateResults}
\end{figure}

\subsection{Refinement}

The outputs of our HDR reconstruction networks for shading and albedo can be combined using the intrinsic model to give us an initial estimation for the HDR image:
\begin{equation}
    \hat{I}_H = A_H * S_H, \quad S_H = \frac{1 - D_H}{D_H}.
\label{eq:inferredHDR}
\end{equation}
However, as the two intrinsic components are estimated separately, the perfect reconstruction of the HDR image is not guaranteed as shown in Figure~\ref{fig:PipelineIntermediateResults}.
In the final step of our pipeline, we train a refinement network that takes the intrinsic HDR reconstruction results as input and generates our final HDR reconstruction result. 

This network has a similar goal to the hallucination network used in Liu~\etal~\cite{liu2020single}: Reconstructing the final HDR image by extending dynamic range. 
However, as it receives the intrinsic reconstruction of the HDR image readily as input, its task is significantly easier. Unlike a network trained for direct LDR-to-HDR conversion,
it does not need to reason about the high-level cues in the image which is performed by our HDR shading reconstruction network. 
Similarly, it does not need to recover the corrupted colors in saturated regions, which is performed by our HDR albedo reconstruction network. 
It is simply tasked with combining the rich input information it receives to reconstruct the final HDR image, which allows it to generate more accurate and sharper results.

Following the inverse representation we adopt for shading, we use the inverse representation for the HDR images as a bounded representation that fits the long-tailed distribution in HDR images between $[0,1]$:
\begin{equation}
    J_H = \frac{1}{I_H + 1}, \quad \hat{J}_H = \frac{1}{\hat{I}_H + 1}, \quad J^*_H = \frac{1}{I^*_H + 1},
\end{equation}
where $J_H$, and $J^*_H$ denote the estimated and ground-truth inverse HDR images and $\hat{J}_H$ is the inferred inverse HDR image following Eq.~\ref{eq:inferredHDR}.

Our refinement network receives the LDR image $I_L$, the inferred inverse HDR image $\hat{J}_H$, the estimated HDR inverse shading $D_L$, and the estimated HDR albedo $A_L$ as a concatenated $h \times w \times 10$ input.
We use the MSE and multi-scale gradient losses to train our refinement network:
\begin{equation}
    \mathcal{L}_{mse}^{J} = MSE \left( J^*_H, J_H \right), \quad
    \mathcal{L}_{msg}^J = MSG \left( J^*_H, J_H \right),
\end{equation}
with the final loss defined as $ \mathcal{L}^J = \mathcal{L}_{mse}^{J} + \gamma \cdot \mathcal{L}_{msg}^J$, $\gamma=0.1$.

\subsection{Network architecture and training}

We utilize the same architecture for our 3 networks. 
We use EfficientNet~\cite{tan2019efficientnet} as the backbone with the decoder from~\cite{midas}, replacing the output activation with a \texttt{sigmoid}. We use RAdam~\cite{radam} as the optimizer with a learning rate of $1e^{-4}$ and CosineAnnealing~\cite{Loshchilov2016SGDRSG} scheduling.

To train our intrinsic formulation, we use the provided synthetic intrinsic ground truth from Hypersim~\cite{roberts:2021} and construct ground truth intrinsics from the raw images of the MultiIllum dataset~\cite{murmann19} as described in~\cite{careaga2023intrinsic}. We follow the preprocessing steps from Liu~\etal~\cite{liu2020single} to create clipped, linearized LDR images with randomly sampled exposure values $t \in [-3..3]$ from the ground truth data which we then decompose into our LDR input intrinsics using \cite{careaga2023intrinsic}.

Our refinement network in contrast does not require supervision in intrinsic space. We can thus extend our training set by real images from MultiRAW~\cite{li2022efficient}, LSMI~\cite{kim2021large}, and the HDR-Real training subset from Liu~\etal~\cite{liu2020single}. For this stage, we follow the ISP model in~\cite{liu2020single} to convert the ground truth images into JPEG files, use their pre-trained networks to retrieve linearized LDR images, and again retrieve the intrinsic decomposition via~\cite{careaga2023intrinsic}. We provide a more detailed description of our training procedure in the supplementary material.

\section{Experiments}
\label{sec:experiments}

In this section, we present a thorough evaluation of our proposed method using multiple standard benchmarks. First, we discuss the datasets, metrics and prior work used in our experimental setup, then we present qualitative and quantitative comparisons to state-of-the-art approaches. 
We present a more detailed evaluation and our ablation studies in the supplementary material.

\subsection{Quantitative evaluation}
\label{sec:experiments:quantitative}
To quantitatively evaluate our approach against existing methods we follow the evaluation protocol proposed by Hanji~\etal~\cite{hanjiComparisonSingleImage2022}. For a given ground truth HDR image, we follow the preprocessing of Liu~\etal~\cite{liu2020single} and map the values into the range $[1-1000]$. We determine the best scale to align the input reconstruction with the ground truth with least-squares, using only values between the $10^{th}$ and $90^{th}$ percentiles to avoid outliers. We then perform CRF correction to match the reconstruction to the ground truth by fitting a per-channel third-degree polynomial in log-RGB space. 

\paragraph{Metrics}
We follow the guidance of Hanji~\etal~\cite{hanjiComparisonSingleImage2022} when choosing metrics to evaluate our approach. In their study, they find that certain metrics better correlate with results from qualitative experiments. We focus our evaluation on the popular HDR-VDP3~\cite{vdp3mantiuk2023hdr} metric as it is purpose-built for HDR evaluation and models the subjective evaluation of a human observer. We specifically use the "quality" task of HDR-VDP3. Furthermore, we evaluate multiple recommended SDR metrics using the PU21~\cite{azimi2021pu21} encoding and provide those results in the supplementary material. 

\paragraph{Datasets}
We evaluate our method on five publicly available datasets. 
We use the recent high-resolution SI-HDR Benchmark~\cite{hanjiComparisonSingleImage2022} as well as commonly used real-world datasets HDR-Eye~\cite{hdreye}, HDR-Real~\cite{liu2020single}, HDR-Synth~\cite{liu2020single}, and RAISE~\cite{Dang2015raise}.

We compare our method against a collection of baselines that share their implementation publicly. 
Table \ref{tab:main_quantitative} shows the results of our quantitative evaluation. 
Each value represents the average HDR-VDP3 score in Just-Objectionable-Differences (JOD) units. 
We show a significant improvement in quality in the SI-HDR Benchmark~\cite{hanjiComparisonSingleImage2022} while achieving state-of-the-art performance on 3 of the other 4 datasets. 
As our qualitative evaluation in the next section and our results presented in the supplementary material demonstrate, the favorable performance of our method comes from our physically-motivated intrinsic formulation, allowing us to reliably reconstruct the high luminance details and recover lost colors. We provide additional metrics and analysis in the supplementary material.

\begin{table}[t]
\centering
\caption{Quantitative results against state-of-the-art using the HDR-VDP3 metric. Our method achieves state-of-the-art performance on the difficult high-definition SI-HDR benchmark dataset. We are state-of-the-art or competitive on all other datasets.}
\begin{tabular}{l|ccccc}
\toprule
     Method &     SIHDR &   HDR-EYE &  HDR-REAL &  HDR-SYNTH &     RAISE \\
\midrule
       Ours                                            &   \topscore{8.96} &  \topscore{8.96} &   \topscore{7.57} &   8.21             &  \underline{8.89} \\
      DrTMO~\cite{endoDeepReverseTone2017}             &   8.27            &  8.72            &   \topscore{7.57} &   \underline{8.27} &              8.64 \\
     HDR-CNN~\cite{eilertsenHDRImageReconstruction2017} &   8.39            &  8.61            &   6.70            &   8.03             &              8.47 \\
  ExpandNet~\cite{marnerides2018expandnet}             &   8.67            &  8.49            &   6.97            &   7.69             &              8.27 \\
  Single-HDR~\cite{liu2020single}                       &   8.79            &  \topscore{8.96} &   \underline{7.47}            &   \topscore{8.51}  &   \topscore{8.95} \\
    Mask-HDR~\cite{santosSingleImageHDR2020}            &   8.25            &  8.60            &   7.10            &   8.14             &              8.44 \\
    HDRUNet~\cite{chen2021hdrunet}                     &   \underline{8.82}            &  8.46            &   5.71            &   7.61             &              8.04 \\
       Multi-Exp Gen.~\cite{le2023single}                     &   8.64            &  \underline{8.87}            &   7.34            &   8.16             &              8.66 \\
       Lightweight~\cite{guo2022lhdr}                         &   8.26            &  8.20            &   7.03            &   7.31             &              7.68 \\
\bottomrule
\end{tabular}
\label{tab:main_quantitative}
\end{table}

\begin{figure}[t!]
    \centering
    \includegraphics[width=\linewidth]{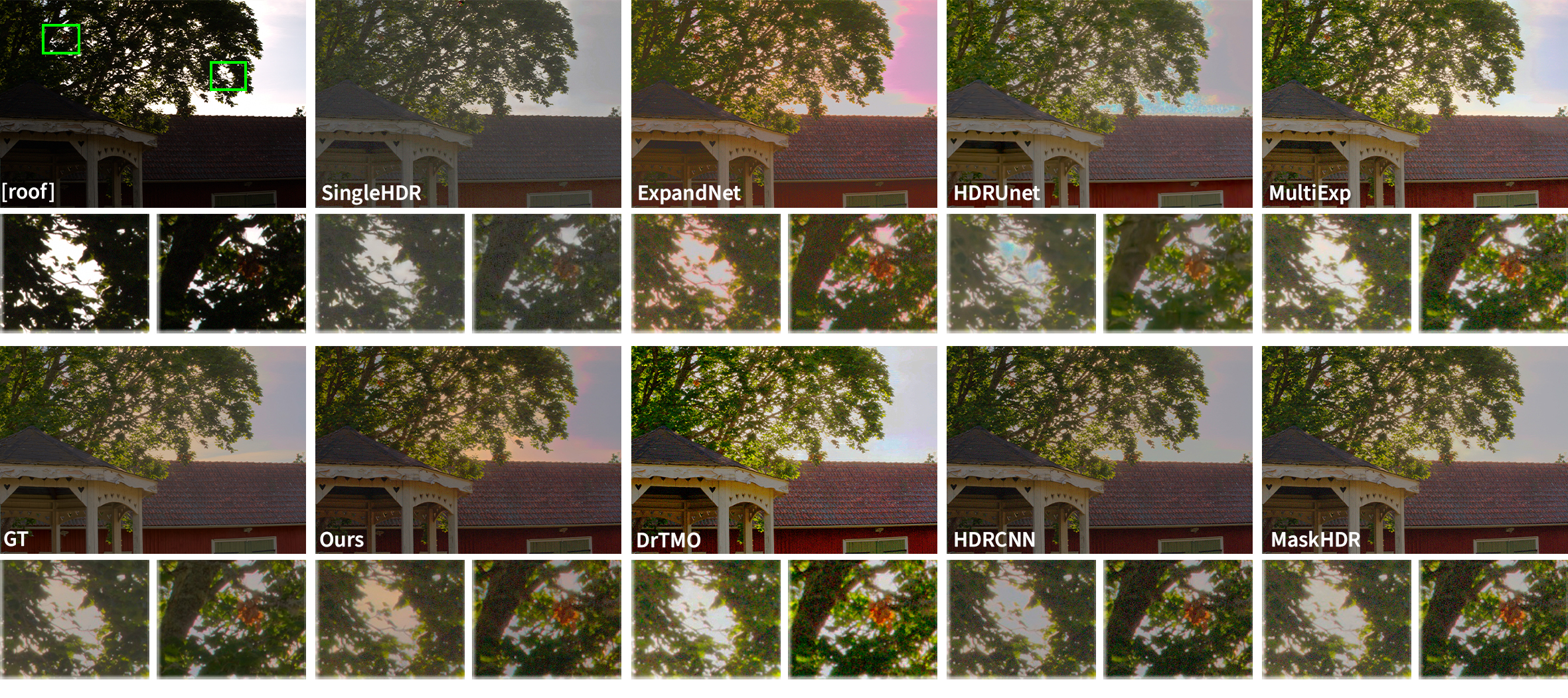}
    \includegraphics[width=\linewidth]{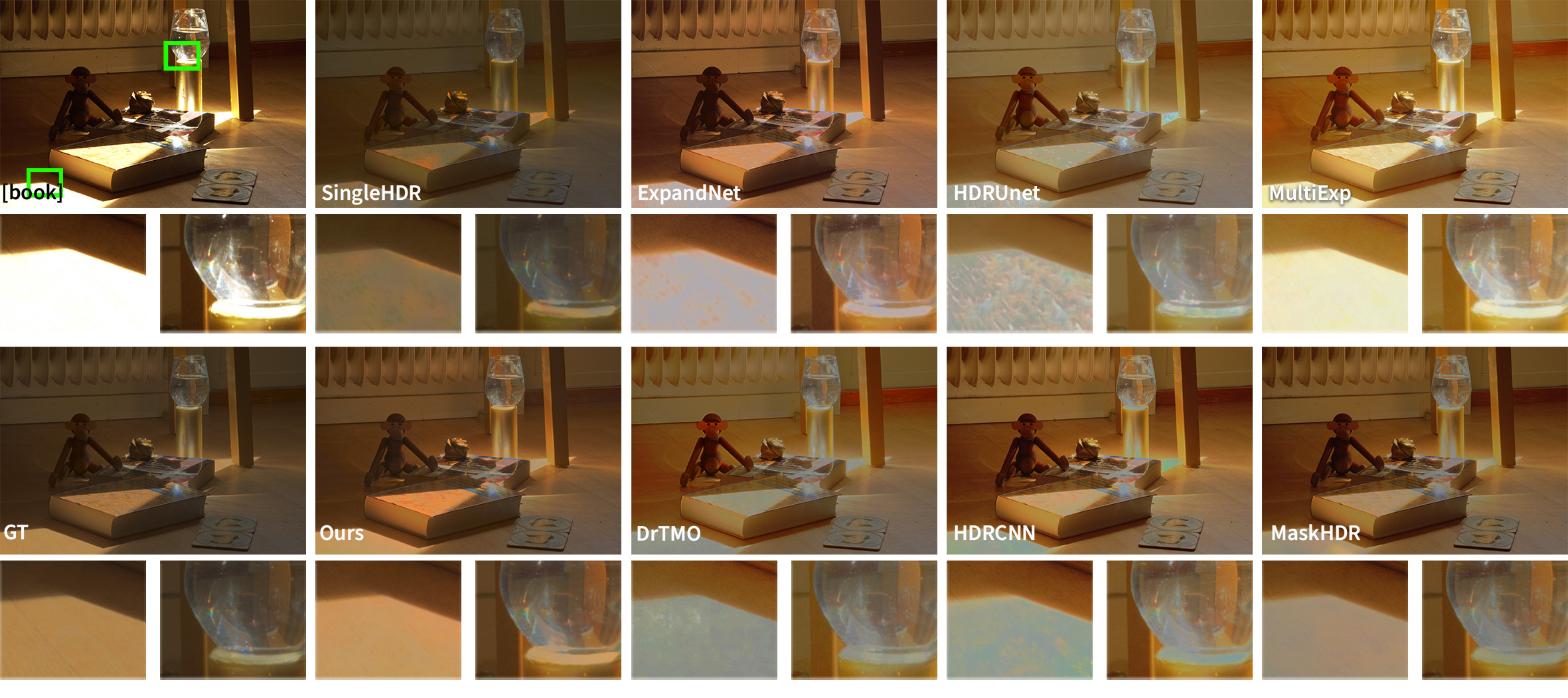}
    \includegraphics[width=\linewidth]{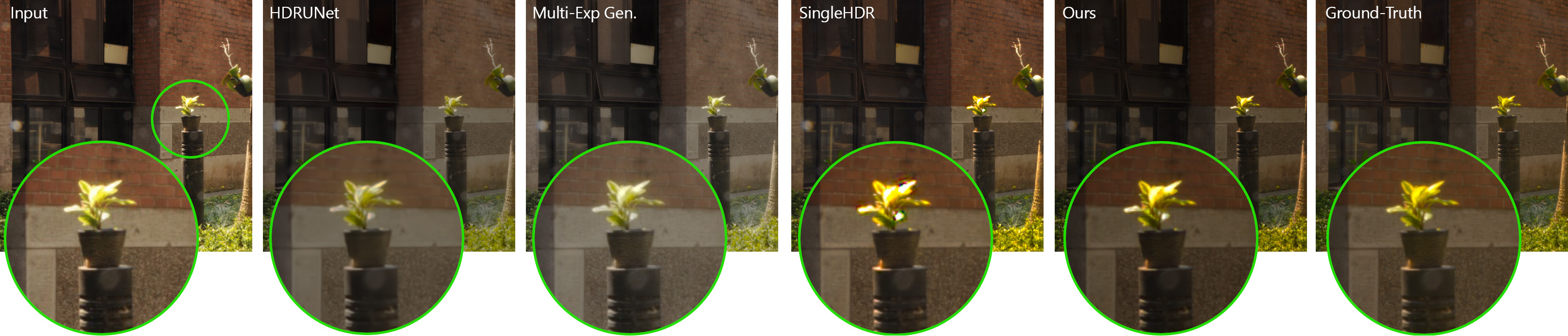}
     \caption{We show results from our method on SI-HDR~\cite{hanjiComparisonSingleImage2022} to the other baselines. We refer to~\cref{sec:experiments:qualitative} and the supplementary material for an in-depth discussion.}
    \label{fig:experimental_all}
\end{figure}

\subsection{Qualitative evaluation}
\label{sec:experiments:qualitative}
\begin{figure}[t]
    \centering
    \includegraphics[width=\linewidth]{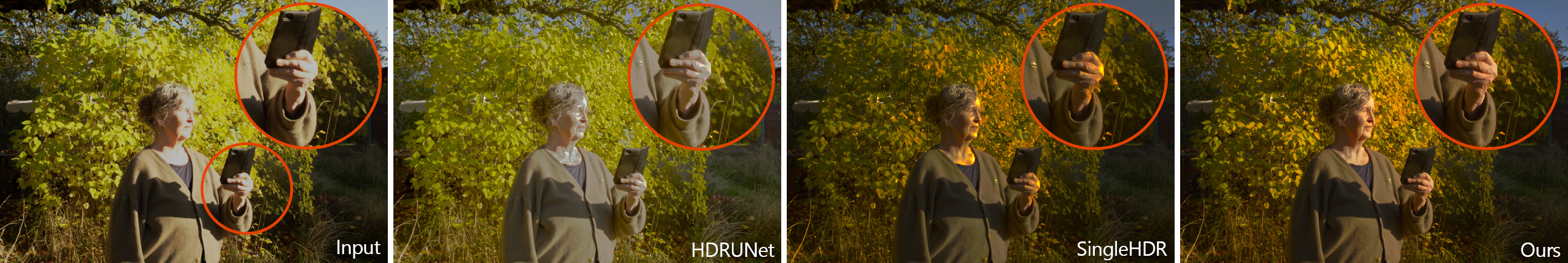}
    \includegraphics[width=\linewidth]{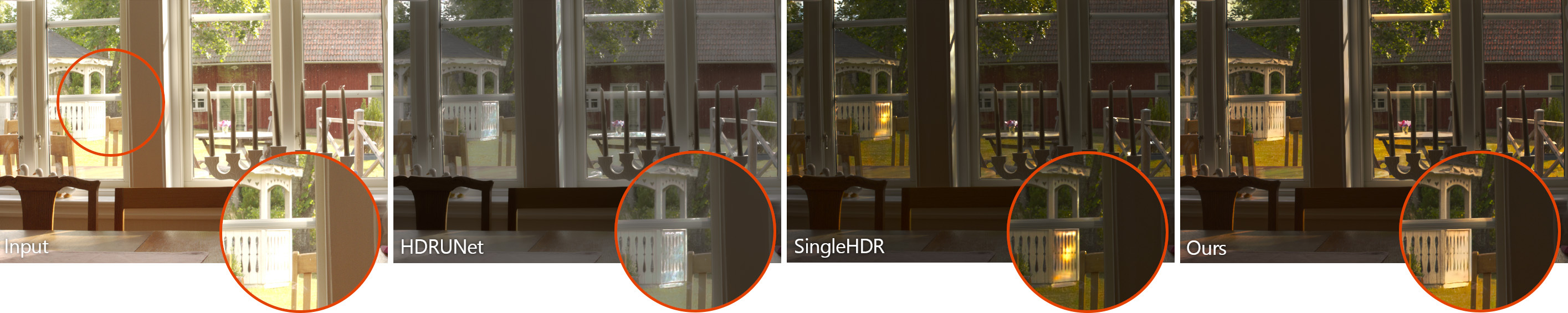}
    \includegraphics[width=\linewidth]{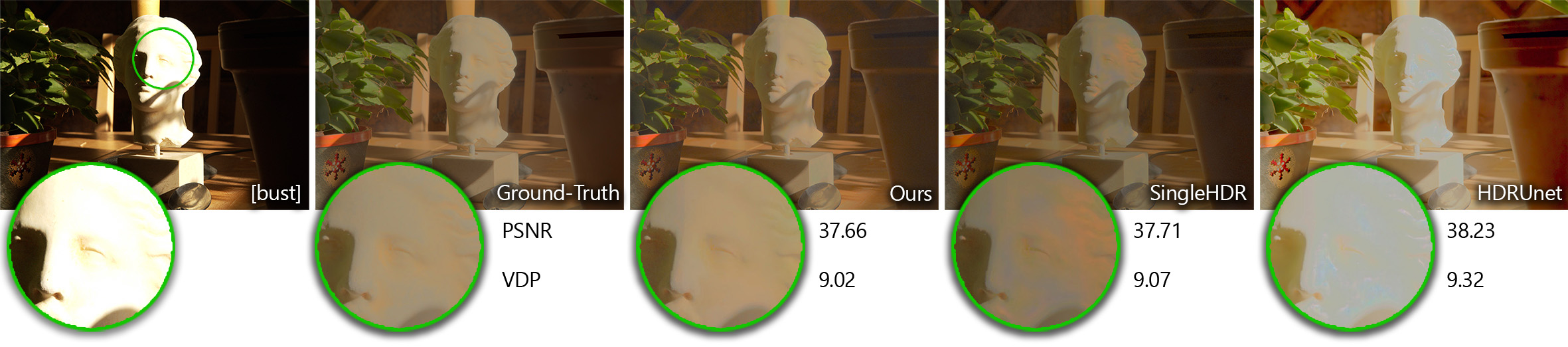}
    \includegraphics[width=\linewidth]{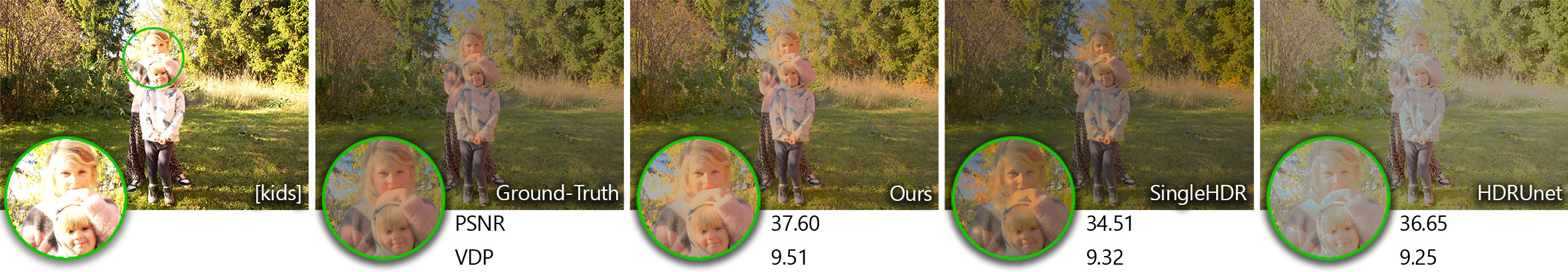}
    \caption{We compare results from our method with a focus on color reconstruction against HDRUnet~\cite{chen2021hdrunet} and SingleHDR~\cite{liu2020single}.We refer to~\cref{sec:experiments:qualitative} and the supplementary material for an in-depth discussion.}
    \label{fig:experimental_singlehdr}
\end{figure}

We present qualitative comparisons in Figures~\ref{fig:experimental_all} and \ref{fig:experimental_singlehdr}, showing the differences in both detail and color reconstruction. On SI-HDR~\cite{hanjiComparisonSingleImage2022}, prior works struggle with predicting accurate details around edges where over- and under-exposed pixels meet. This becomes notable around the leaves in the [roof] scene with visible halos and color shifts. 
On the floor in the [book] scene, most competitors fail to reconstruct the homogenous surface and hallucinate patterns and color blobs. In both cases, our method seamlessly recovers missing details in over-exposed areas without artifacts. For the woman in~\cref{fig:experimental_singlehdr}, HDRUnet~\cite{chen2021hdrunet} and SingleHDR~\cite{liu2020single} both exhibit strong artifacts as bright spots in the hand and the face, similar to the faces of the [bust] and the [kids]. Our method in contrast can recover the correct color information due to our modeling of the albedo.
The results showcase the advantage of the intrinsic formulation for creating visually pleasing results. We provide an extensive qualitative comparison in the supplementary material.

\section{Conclusion}
\label{sec:discussion}

In this work, we remodel the HDR reconstruction problem in the intrinsic domain and propose a multi-stage pipeline to recover lost details. 
The luminance range extension can be conveniently modeled in the shading space which allows the network to better model the correlation between illumination and geometry. 
Similarly, neural networks can easily take advantage of albedo sparsity to recover lost color information with high fidelity. 
Through qualitative and quantitative evaluations, we show that our physically-inspired modeling of the HDR reconstruction problem achieves state-of-the-art performance, improving upon prior work in terms of both detail recovery and color reproduction in a wide variety of in-the-wild photographs.

\section*{Acknowledgements}
We acknowledge the support of the Natural Sciences and Engineering Research Council of Canada (NSERC), [RGPIN-2020-05375].

\bibliographystyle{splncs04}
\bibliography{hdr_bib}
\end{document}